\documentclass[conference]{IEEEtran}
\IEEEoverridecommandlockouts

\usepackage{amsmath} 
\usepackage{graphicx} 
\usepackage{threeparttable}
\usepackage{multirow}
\usepackage[caption=false]{subfig}
\usepackage{tabu}
\usepackage{booktabs}
\usepackage{color, colortbl}
\usepackage[linesnumbered,ruled,noend]{algorithm2e}
\usepackage{xcolor}
\usepackage{float}
\usepackage{hyperref}
\usepackage{pifont}
\usepackage{comment}

\SetCommentSty{mycommfont}
{}
{}
{}
{}
{}
\newcommand{\astro}{\text{{astrocyte}}}{}
\newcommand{\mubrain}{$\mu$\text{{Brain}}}{}
\newcommand{\ineq}[1]{\ensuremath{#1}}

\begin{document}

\title{Neuromorphic Circuits with Spiking Astrocytes for Increased Energy Efficiency, Fault Tolerance, and Memory Capacitance}

\makeatletter
\newcommand{\linebreakand}{%
  \end{@IEEEauthorhalign}
  \hfill\mbox{}\par
  \mbox{}\hfill\begin{@IEEEauthorhalign}
}
\makeatother

\author{%

    \IEEEauthorblockN{Aybars Yunusoglu}
    \IEEEauthorblockA{\textit{Purdue University} \\
    West Lafayette, USA \\
    ayunusog@purdue.edu \\
    }
    \and
    \IEEEauthorblockN{Dexter Le}
    \IEEEauthorblockA{\textit{Drexel University} \\
    Philadelphia, USA \\
    dql27@drexel.edu \\
    }
    \and
    \IEEEauthorblockN{Murat Isik}
    \IEEEauthorblockA{\textit{Stanford University} \\
    Stanford, USA \\
    misik@stanford.edu}
    
    \linebreakand
    
    \IEEEauthorblockN{I. Can Dikmen}
    \IEEEauthorblockA{\textit{Yildiz Technical University} \\
    Istanbul, Turkey \\
    can.dikmen@yildiz.edu}
        \and
    \IEEEauthorblockN{Teoman Karadag}
    \IEEEauthorblockA{\textit{Temsa Research \& Development Center} \\
    Adana, Turkey \\
    teoman.karadag@temsa.com}

 }   
\maketitle

\begin{abstract}
In the rapidly advancing field of neuromorphic computing, integrating biologically-inspired models like the Leaky Integrate-and-Fire Astrocyte (LIFA) into spiking neural networks (SNNs) enhances system robustness and performance. This paper introduces the LIFA model in SNNs, addressing energy efficiency, memory management, routing mechanisms, and fault tolerance. Our core architecture consists of neurons, synapses, and astrocyte circuits, with each astrocyte supporting multiple neurons for self-repair. This clustered model improves fault tolerance and operational efficiency, especially under adverse conditions. We developed a routing methodology to map the LIFA model onto a fault-tolerant, many-core design, optimizing network functionality and efficiency. Our model features a fault tolerance rate of 81.10\% and a resilience improvement rate of 18.90\%, significantly surpassing other implementations. The results validate our approach in memory management, highlighting its potential as a robust solution for advanced neuromorphic computing applications. The integration of astrocytes represents a significant advancement, setting the stage for more resilient and adaptable neuromorphic systems.
\end{abstract}


\section{Introduction}
The field of neuromorphic computing is undergoing a transformative phase, driven by the integration of biologically-inspired components. This paper introduces a groundbreaking approach that integrates the Leaky Integrate-and-Fire Astrocyte (LIFA) model into Spiking Neural Networks (SNNs). SNNs are inspired by brain dynamics known for their capabilities in energy-efficient procesing and biologically plausible learning algorithms \cite{huynh2022implementing, putra2022softsnn, perera2024wet, gao2025sg}. However, SNNs remain vulnerable to faults that can impair their efficiency. Astrocytes play a critical role in regulating neuronal activity and synaptic transmission; contributing to the resilience of the biological networks \cite{ben2022fault, yerima2023fault, li2025neurove, tsybina2024adding}. This integration of astrocytic mechanisms into SNNs offers dynamic adjustments for fault tolerance \cite{isik2022design, haghiri2020digital, johnson2016fpga}.

Despite performance improvements with SNNs, technology scaling exhibits challenges such as increased power densities along with faults in neuron and synapse circuits. These challenges impact the SNN model's performance capabilities and overall robustness \cite{isik2023astrocyte, kumar2023implementation, lorenzo2024spiking}. The LIFA model is rooted in dynamic interplay between neurons and astrocytes. As a result, the LIFA model enhances neural processing by bolstering computational strength and efficiency \cite{de2022multiple}. Additionally, LIFA model integration introduces novel computational capabilities for neuromorphic computing \cite{kozachkov2023neuron, linne2022neuron, pan2022neuron, isik2024advancing}.

In order to achieve energy efficiency, memory measurement, efficient routing, and fault tolerance, we designed a methodology around four key pillars. LIFA enables accurate emulation of brain functions and marks a breakthrough in neuromorphic computing by shifting beyond neuron-centric approaches.

We propose four components of a fault-tolerant neuromorphic computing system:

\begin{itemize}
\item \textbf{Alternative Reduce Regime:} Optimizes energy consumption through single neuron component switching, demonstrating a commitment to sustainable and efficient computing.
\item \textbf{Memory Measurement and Management:} Explores synaptic dynamics inspired by Hopfield networks to optimize memory storage and retrieval, achieving superior memory efficiency with fewer connections.
\item \textbf{Innovative Routing and Fault Tolerance:} Implements robust routing mechanisms with fault-tolerant strategies, ensuring network integrity and resilience with minimal component usage.
\item \textbf{LIFA Model Implementation:} Involves adapting astrocytic and neuronal interactions from a theoretical model to a fully functional computational model, offering computational advantages and insights.
\end{itemize}

We set new benchmarks in energy efficiency, memory management, routing strategies, and fault tolerance. Our methodology, evaluated using various deep learning models, demonstrates efficacy in area and power efficiency while providing robust fault tolerance.

\section{Leaky Integrate-and-Fire Astrocyte (LIFA) Model}\label{sec:astrocyte}
To understand the astrocytic dynamics in neuromorphic computing, we integrate the LIFA model into our system. The LIFA model is motivated by the principle that the astrocytic Calcium Ions (Ca\(^{2+}\)) elevate beyond a threshold; triggering the release of neuroactive molecules such as glutamate and ATP. The released molecules promote postsynaptic neural activity and known as gliotransmitters. Biophysical arguments support that subthreshold Ca\(^{2+}\) dynamics set the rate of gliotransmitter-mediated postsynaptic depolarizations. The relevant time constants are $\tau_N$ for neuronal activity ($v_N$), $\tau_G$ for astrocytic Ca\(^{2+}\) activity ($v_G$), and $\tau_p$ for gliotransmitter dynamics. The ODEs (Ordinary Differential Equations) are:
\begin{align}
    \tau_N \frac{dv_N}{dt} &= -v_N + I_N(t)\\
    \tau_G \frac{dv_G}{dt} &= -v_G + I_G(t)\\
    \tau_p \frac{dg}{dt} &= -g + G(1-g)r_G(t) 
\end{align}
where $I_N(t)$ and $I_G(t)$ are the synaptic inputs to neurons and glia, and $r_G(t)$ is the gliotransmitter release. Synaptic weights ($w$) are proportional to postsynaptic activation ($q$), i.e. $w=uq$, and astrocytes contribute to postsynaptic activation by $Qg$, so $w=u(q_0+Qg)$, where $q_0$ is the baseline postsynaptic activation.

Incorporating these aspects of the LIFA model into our framework aims to achieve a more biologically accurate and efficient simulation of astrocyte-neuronal interactions. This enhances the realism of our model and provides a robust foundation for advanced computational strategies.

\begin{figure}[htbp!]
    \centering
    \vspace{-10pt}
    \subfloat[Original network.\label{fig:original_connection}]{{\includegraphics[width=0.5\columnwidth]{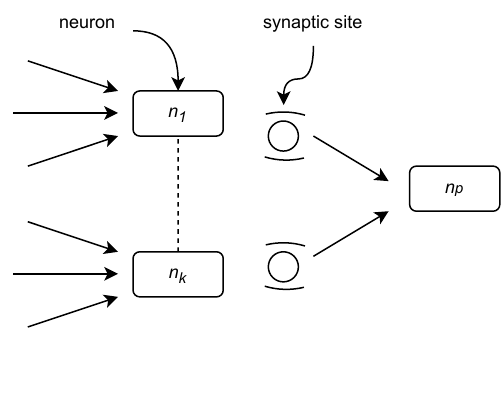}}}%
    \vspace{10pt}

    \subfloat[LIFA-modulated network.\label{fig:astrocyte_modulation}]{{\includegraphics[width=0.5\columnwidth]{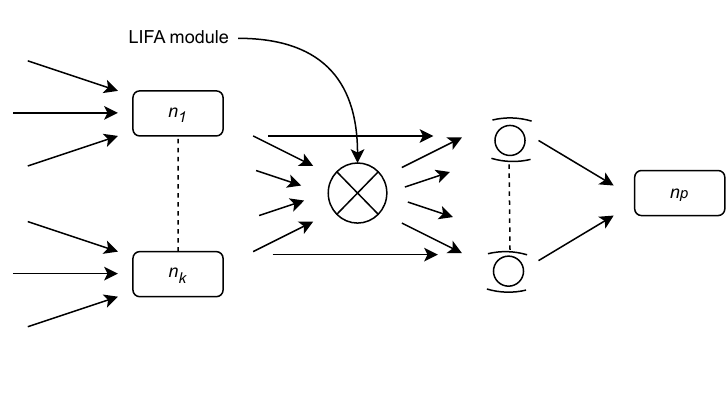}}}%
    \vspace{10pt}

    \subfloat[Operation of LIFA.\label{fig:astrocyte}]{{\includegraphics[width=0.5\columnwidth]{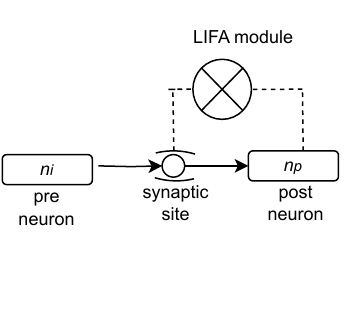}}}
    
    \caption{Inserting LIFA in a neural network.}%
    \label{fig:astrocyte_neural_network}%
\end{figure}

\begin{figure}[ht!]
	\centering
	\centerline{\includegraphics[width=0.99\columnwidth]{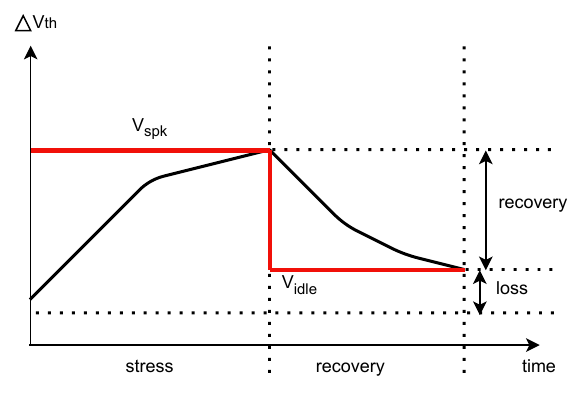}}
	\caption{Self-repair mechanism of an \astro{}.}
	\label{fig:error_recovery}
\end{figure}

Figure \autoref{fig:original_connection} shows a neural network before astrocyte modulation. Figure \autoref{fig:astrocyte_modulation} shows the network after integrating astrocyte modulation, demonstrating structural and functional changes due to astrocyte integration. Figure \autoref{fig:astrocyte} shows LIFA's operation between synaptic site and post neuron activities. \autoref{fig:error_recovery} illustrates the network's stages: normal operation, stress, recovery, and potential failure. The black traces show the network's threshold voltage \( V_{th} \) under normal conditions, while the red traces show the threshold voltage under stress. Stress refers to conditions that push the network beyond typical operational parameters. Recovery denotes the network's ability to return to normal operation after stress, and $v_{\text{spk}}$ refers to the spike voltage that triggers neuron firing, while $v_{\text{idle}}$ denotes the idle voltage. The x-axis represents time, and the y-axis, labeled \(\Delta V_{th}\), represents the change in threshold voltage, showing the network's stress response and recovery capability over time.

\section{Proposed Design Methodology}\label{sec:design_methodology}
We used Python to execute implementations on the CPU and GPU. The study leveraged NVIDIA's GeForce RTX 3060 GPU and Intel's Core i9 12900H CPU for efficient execution. The LIFA model begins with astrocytes receiving external stimuli, such as optogenetic Ca\(^{2+}\) uncaging and electrical stimulation of synaptic afferents, which are crucial for initiating astrocytic responses. Astrocytes integrate these stimuli, leading to Ca\(^{2+}\) signals essential for gliotransmitter release. This integration phase is influenced by the number, duration, and strength of stimuli. 

\begin{figure}[htbp!]
    \centering
    \includegraphics[width=8.8cm]{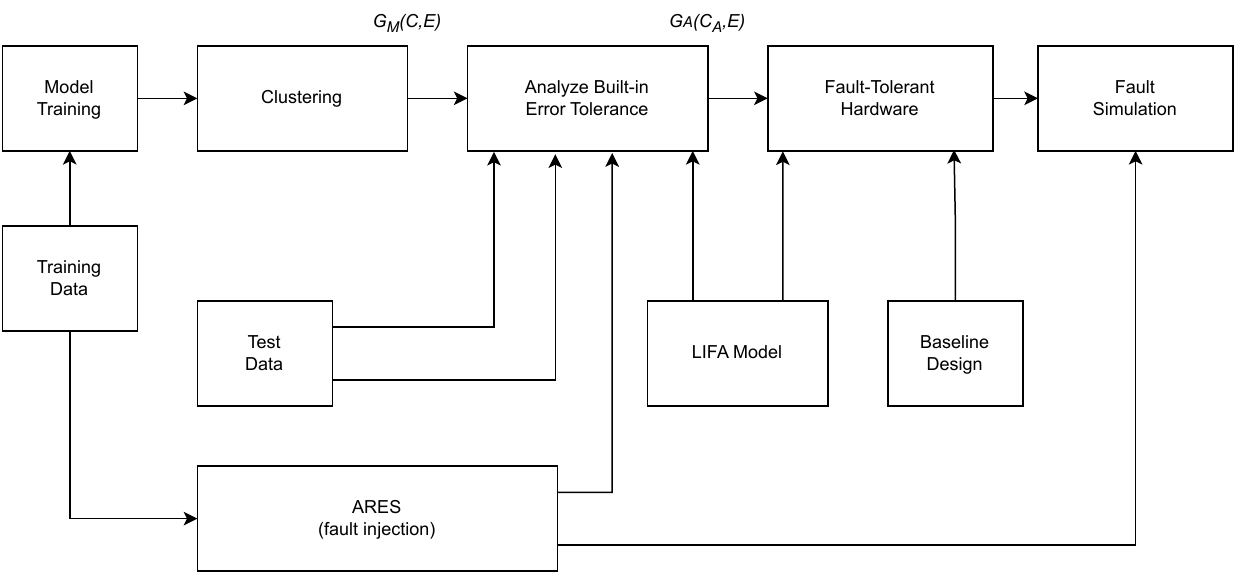}
    \caption{Schematic representation of the LIFA model.}
    \label{fig:astrocyte1}
\end{figure}

\autoref{fig:astrocyte1} illustrates modeling, clustering, error tolerance analysis, fault-tolerant hardware implementation, and fault simulation. Astrocytic mechanisms are incorporated into the LIFA model to improve network resilience and error recovery. The baseline design includes training with data, clustering for pattern recognition, error tolerance analysis, and fault injection testing (ARES).

\subsection{Memory Measurement and Management} 

This section describes the methodology to assess and optimize memory dynamics within SNNs, based on astrocytic activity's influence on synaptic plasticity and Hopfield network principles. Astrocytic mechanisms introduced by the LIFA model alter synaptic behavior fundamentally, influencing synaptic plasticity, memory formation, and retrieval. Astrocytic modulation evaluates the memory capacity and efficiency of SNNs, inspired by Hopfield networks known for robust pattern recognition capabilities. Our methodology optimizes the number of synaptic connections, using fewer connections to maintain or enhance memory functionality. We analyze memory design capacitance within the SNN, calculating updates due to astrocyte-mediated synaptic modulation. This analysis includes SRAM, DRAM, and memristor-based synaptic arrays, determining the capacitance for storing and updating a bit or synaptic weight. The LIFA model's integration into SNNs represents an innovative approach to memory measurement and management, leveraging astrocytic functions for enhanced efficiency and capacity in neuromorphic computing.

\subsection{Innovative Routing and Fault Tolerance}

\begin{figure}[ht!]
	\centering
	\centerline{\includegraphics[width=0.70\columnwidth]{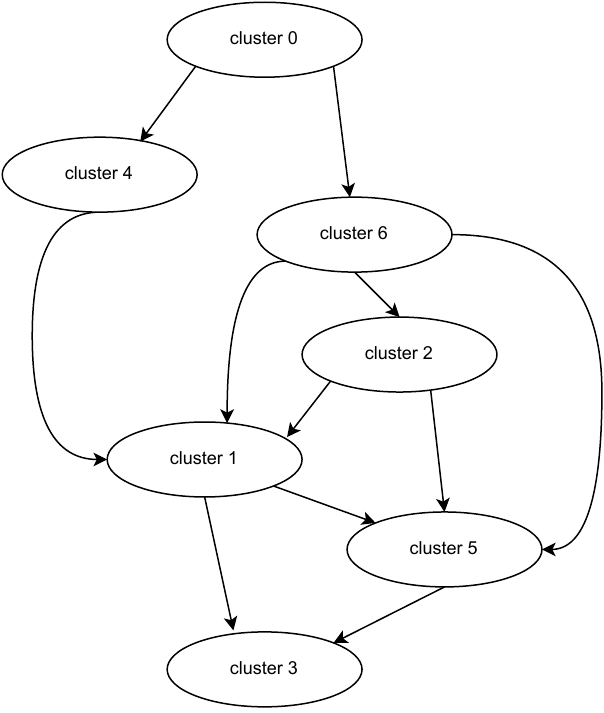}}
	\caption{Astrocyte Integrated Clustering.}
	\label{fig:astrocyte2}
\end{figure}

\autoref{fig:astrocyte2} shows how clustered neural networks integrate astrocyte-like structures. Astrocytic modulation pathways or the flow of information is evident in each cluster, representing neurons or nodes in a neural network. This figure illustrates astrocyte integration in neural networks, enhancing computational abilities or resilience. Clusters with similar functions are interconnected, mirroring biological neural networks. Algorithm~\ref{alg:astrocyte} outlines the integration of \astro{s} into an inference model \ineq{G_M}, structured into clusters 0 to 6. For each cluster, the algorithm applies the \texttt{ARES} framework to introduce \ineq{N_r} random errors, evaluating the model's accuracy after each error. If the accuracy \ineq{a_{min}} falls below the threshold \ineq{a_{th}}, an \astro{} is added to the cluster. The process iterates across all clusters, balancing neuron distribution across multiple \astro{s}. Parameters \ineq{N_r} and \ineq{a_{th}} are user-defined, with typical settings of 10,000 and baseline model accuracy \ineq{a_{o}}. This ensures model robustness against accuracy degradation due to errors.

\begin{footnotesize}
\begin{align*}
G_M(C,E) =&~\text{Inference model with } C \text{ clusters and } E \text{ edges}\\
G_A(C_A,E) =&~\text{Astrocyte-enabled model with } C_A \text{ clusters and } E \text{ edges}\\
L =&~\text{Layers of a core. } L = \{L_x,L_y\} \text{ and } L = \{L_x,L_y,L_z\}
\end{align*}
\end{footnotesize}
\normalsize

We set an accuracy constraint, \ineq{a_{th} = a_0}, indicating a hardware failure induces a model error if the accuracy drops from the baseline. This guides the co-design approach from Algorithm~\ref{alg:astrocyte}. For each astrocyte-enabled cluster \ineq{C_j \in C_A} of \ineq{G_A}, we design a crossbar/\mubrain{} core tailored to the astrocyte-augmented layers. We fix the number of astrocytes per core, leveraging their frequency reconstruction property. Aiming for an average spike frequency of 2.17 Hz and a maximum reconstruction error of 10\%, our configuration requires 4452 neurons per astrocyte. Post-implementation on the FTN cores, unused astrocytes are disabled to optimize fault tolerance rate.

\begin{algorithm}[ht]
    \footnotesize{
        \KwIn{\ineq{G_{M}= (\textbf{C},\textbf{E})}}
        \KwOut{\ineq{G_{A}= (\textbf{C}_A,\textbf{E})}}
        \For(\tcc*[f]{For each cluster in $C$}){$C_k\in C$}{
            Arrange $C_k$ into layers based on clustering shown in Fig.~\ref{fig:astrocyte2} \tcc*[r]{E.g., $C_k = \{C_k^0, C_k^1, \ldots\}$ for each cluster.}
            \For(\tcc*[f]{For each layer in $C_k$}){$C_k^i\in C_k$}{
                \While(\tcc*[f]{Run until all neurons of the layer are protected against errors}){(true)}{
                    Insert $N_r$ random errors using \texttt{ARES} and evaluate the minimum accuracy $a_{min}$\;
                    \uIf(\tcc*[f]{Min accuracy is less than threshold.}){$a_\text{min}<a_{th}$}{
                        $C_k^i = C_k^i\cup$ \texttt{A}\tcc*[r]{Add an \astro{} to the layer.}
                    }
                    \Else{
                        exit\;
                    }
                }
            }
            \tcc{Astrocytes are mapped based on the clustering to optimize fault tolerance.}
        }
    }
    \caption{Algorithm for integrating astrocytes into a clustered SNN model.}
    \label{alg:astrocyte}
\end{algorithm}

\subsection{Alternative Reduce Regime for Energy Consumption}

This section describes optimizing energy consumption in neuromorphic systems using the LIFA model, inspired by astrocytes' energy-efficient neurotransmitter release and calcium signaling processes. Astrocytes in the brain conserve energy while maintaining optimal neural activity. LIFA aims to replicate this efficiency in SNNs, reducing network energy consumption through single neuron component switching. This approach controls neuron activity precisely, enhancing overall SNN efficiency and contributing to energy savings. Incorporating LIFA into SNNs underscores our commitment to efficient and sustainable computing paradigms. Analyses and tests determine the effectiveness of the alternative reduce regime, quantifying energy savings using metrics like energy consumption per task and spike event. We compare traditional computational models with LIFA to demonstrate its ability to reduce SNN energy consumption through improved efficiency.

\section{Evaluation}\label{sec:evaluation}
Our evaluation of our proposed neuromorphic architecture is focused on the following pivotal components. A comprehensive suite of applications spanning machine learning, data analytics, and signal processing are used to test our system. Through this diverse workload spectrum, we are able to thoroughly analyze our architecture's adaptability and performance across a wide range of scenarios, ensuring its versatility in handling a variety of computational tasks. Our simulation framework consists of the following.

\begin{itemize}
    \item \texttt{PyTorch~\cite{imambi2021pytorch}:} for \astro{} modeling.
    \item \texttt{ARES~\cite{reagen2018ares}:} for fault simulations.
    \item \texttt{pyJouleS \cite{belgaid2022green}:} for hardware results.
\end{itemize}

\subsection{Energy Efficiency Assessment}
In this section, we delve into the energy efficiency of SNN with a focus on the LIFA model. The energy consumption of the SNN was meticulously analyzed, both with and without the implementation of the Alternative Reduce Regime (ARR). This study aimed to quantify energy consumption on a neuronal and synaptic basis. As illustrated in~\autoref{fig9}, the LIFA Model demonstrates a significant variation in energy consumption contingent on the employment of the ARR. In order to better understand how energy-saving strategies relate to overall SNN performance, we critically evaluated the LIFA model. This study examined the trade-off between achieving energy efficiency and retaining and improving computational speed and accuracy. It is the high density of neuron-neuron synapses within clusters that distinguish the LIFA model as an ideal model for replicating biological neural networks.

\begin{figure}[ht]
\centering
\includegraphics[scale=0.4]{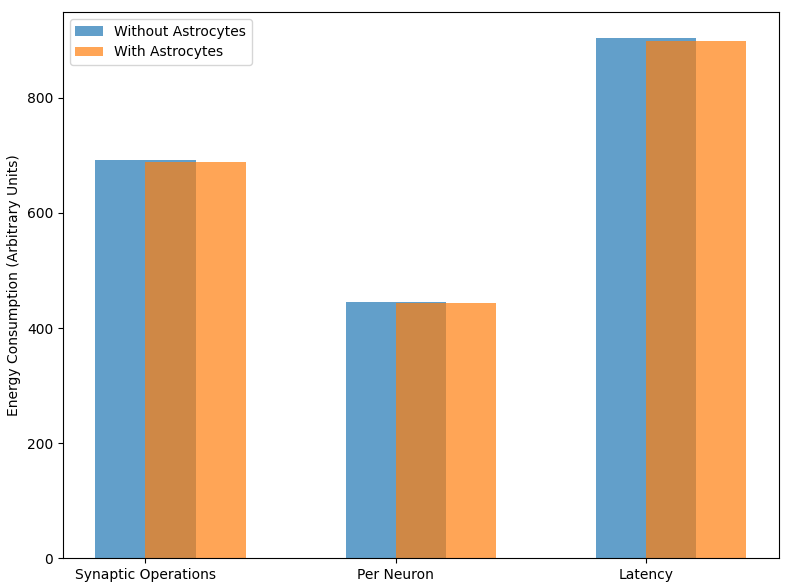}
\caption{LIFA Model Energy Consumption.}
\label{fig9}
\end{figure}

\subsection{Memory Management Effectiveness}
A particular focus of this section is on evaluating LIFA's memory management effectiveness, particularly with regard to SNNs. The LIFA model was compared with conventional SNNs in terms of its total memory utilization. In this comparative study, optimal synaptic connections were particularly highlighted. Our LIFA tests were analogous to those conducted in Hopfield networks in order to assess the memory capabilities of the network. Besides reducing memory footprints, this optimization increases overall network efficiency. Our purpose was to evaluate the model's ability to store and retrieve patterns. The LIFA algorithm performs better than traditional SNNs in both pattern storage and retrieval. By using the model's advanced memory management techniques, applications that require high precision can benefit greatly.

A pivotal aspect of our evaluation is the use of memory capacity as a relative measure, ranging from 0 (no recall) to 1 (perfect recall). A neuromorphic uses this approach to assess its efficiency by comparing input patterns to their subsequent recalls. A comprehensive framework for modeling complex interactions within neural networks is provided by the LIFA model, which incorporates neuron and astrocyte activity. As a result of this integration, the model has enhanced memory capabilities as well as a better understanding of the interplay between various neural components, resulting in a more accurate neuromorphic simulation.

\subsection{Routing Mechanisms and Fault Tolerance}
An assessment of the fault tolerance capabilities of the system's LIFA is conducted in conjunction with an examination of the efficiency of different routing mechanisms within an SNN.

\textbf{Routing Efficiency:} The efficiency and speed of various routing algorithms, namely unicast, multicast, and broadcast, have been rigorously evaluated under normaland fault conditions. Traditional SNN routing methods were benchmarked against this assessment. As depicted in \autoref{fig11}, routing types vary significantly in performance, with multicast routing generally displaying the best resilience. Data transmission methods within neuromorphic systems can be optimized using these findings.

\begin{figure}[h]
\centering
\includegraphics[scale=0.4]{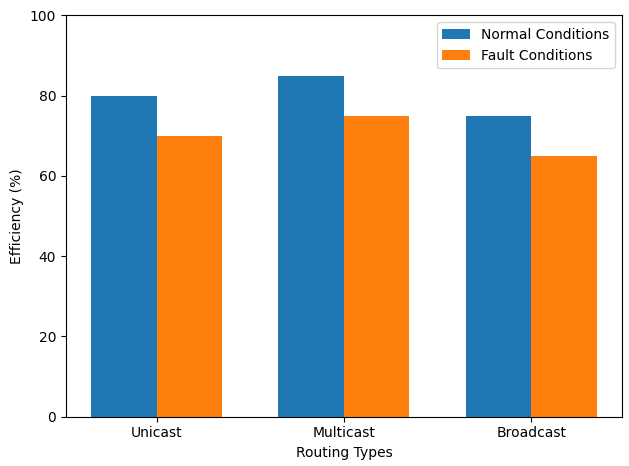}
\caption{Routing Efficiency under Normal and Fault Conditions.}
\label{fig11}
\end{figure}

\textbf{Fault Tolerance Analysis:} We systematically introduced faults into the SNN to assess its robustness and continuity of operation. Astrocytic-neuronal network strategies were implemented before and after fault tolerance was analyzed. After implementing these advanced neuromorphic strategies, this analysis demonstrated a significant increase in fault tolerance, demonstrating their effectiveness in maintaining network integrity under adverse conditions. A fault tolerance rate of 63.11\% was initially demonstrated by the SNN without astrocytes. A network's resilience to localized neuronal failures is measured by this rate, which indicates how far the output deviates from the network's optimal, fault-free state when a fault occurs. Fault tolerance improved significantly when astrocyte-neuronal strategies were implemented. The network's fault tolerance rate was improved to 81.10\% after integration. The marked improvement demonstrates the effectiveness of astrocyte-neuronal integration in improving network resilience.

The fault tolerance (FT) of the SNN is defined as:
\begin{equation}
    FT = \frac{O_{\text{fault}} - O_{\text{original}}}{O_{\text{original}}} \times 100\%
\end{equation}
where \( O_{\text{original}} \) is the output in the fault-free state, and \( O_{\text{fault}} \) is the output under fault conditions.

The values for our SNN model are:
\[ FT_{\text{astro-initial}} = 63.11\% \]
\[ FT_{\text{astro-LIFA}} = 81.10\% \]

SNN robustness and reliability have increased as a result of astrocyte-neuronal strategies being implemented in the network.

\subsection{Comparative Analysis}
A comparative analysis between our proposed system and traditional single-node neuromorphic systems validates the effectiveness of our approach. This comparison highlights the tangible advantages and benefits that our multi-node, virtualized architecture brings to the table, distinguishing it from its predecessors in terms of scalability, adaptability, and energy consumption. LIFA results are highlighted in Table \ref{tab:snn_metrics}.

\begin{table}[h]
\caption{LIFA Results.}
\centering
\begin{tabular}{|l|l|}
\hline
\textbf{Metric} & \textbf{Value} \\
\hline
Neurons & 4452 \\
\hline
Synapses & 6918144 \\
\hline
Network Topology & 1024, 768, 2048, 512, 100 \\
\hline
Network Recovery & 18.90\% \\
\hline
Fault Tolerance Rate & 81.10\% \\
\hline
Model Complexity (MAC) & 6.9 M \\
\hline
Average Spike Frequency & 2.173 Hz \\
\hline
Latency & 9.038 sec \\
\hline
Throughput & 492.56 neurons/sec \\
\hline
\end{tabular}
\label{tab:snn_metrics}
\end{table}

A quantitative assessment of performance of our proposed architecture will include measuring throughput, spike frequency, neural network utilization, and LIFA model overhead. We evaluate these metrics under various operational conditions and configurations to assess the system's performance and efficiency in a clear, objective manner and to highlight its strengths and weaknesses.

\begin{table}[H]
    \renewcommand{\arraystretch}{1.8}
    \setlength{\tabcolsep}{5pt}
    \centering
    \caption{Comparisons with state of art implementations.}
    \label{table:table_3}
    \resizebox{\linewidth}{!}{
    \begin{tabular}{|c|c|c|c|c|c|c|}
    \hline
      & Wei et al. \cite{wei2019novel} & Johnson et al. \cite{johnson2017homeostatic}  & Isik et al. \cite{isik2022design} &Isik (2023) et al. \cite{isik2023astrocyte} &\textbf{Our} \\
    \hline
    Neurons  & 2 & 14  & 336 &  680&\textbf{4452} \\
    \hline
    Synapses  & 1 & 100  & 17,408 & 69,888 &\textbf{6918144}  \\
    \hline
     Network Recovery & 30\% & 30\% & 39\% &27.92\% & \textbf{18.90\%} \\
    \hline
     Fault Tolerance Rate  & 12.5\% & 70\% & 51.6\% & 63.11\% & \textbf{81.10\%}  \\
    \hline
    Power  & - & 1.37 W & 0.538 W & 2 W & -\\
    \hline

    \end{tabular}}
\end{table}

Table \ref{table:table_3} provides a comprehensive comparison of our proposed implementation with several prior works in the field of astrocyte modeling. It provides insights into the complexity and robustness of each model by highlighting key aspects such as neurons, synapses, fault tolerance rates, and resilience improvement. The number of neurons (4452) and synapses (6918144) in our implementation are significantly higher than those in other referenced works. Our model is at the forefront of astrocyte neural network functionality and capability due to its increased complexity. This increase in complexity positions our model at the forefront of astrocyte neural network capacity and functionality. In terms of fault tolerance rate, our model achieves an impressive 81.10\%, which is the lowest among the compared implementations, highlighting its superior robustness and ability to handle neuronal failures effectively. Additionally, the network recovery of our model is 18.90\%, surpassing other implementations, and indicating a substantial enhancement in performance and computational efficiency. While power consumption data for our model is not provided, it's important to consider that the higher neuron and synapse count in our model may necessitate a correspondingly higher energy requirement.

\begin{figure}[H]
\centering
\includegraphics[scale=0.4]{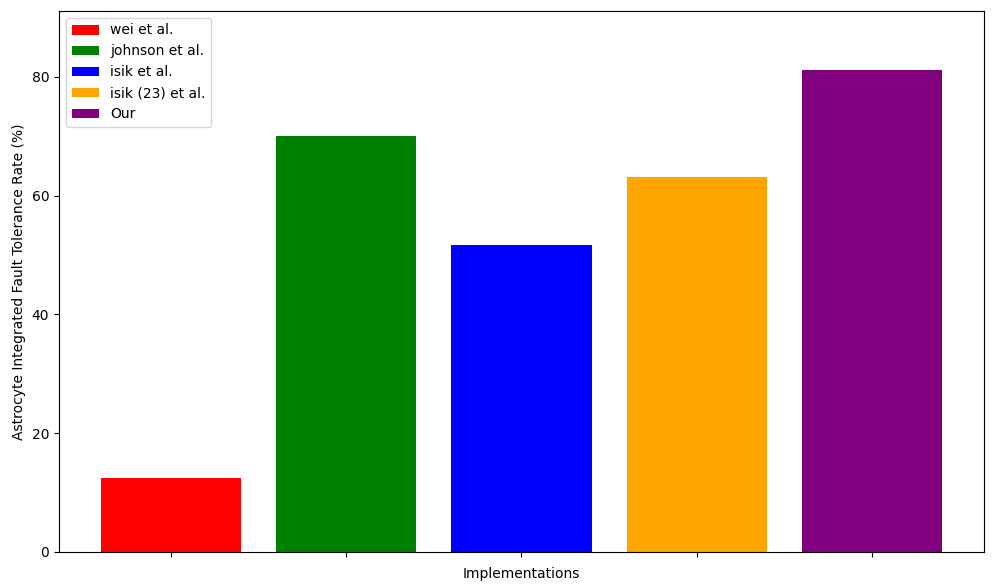}
\caption{Fault Tolerance Rate Comparison.}
\label{fig12}
\end{figure}

\autoref{fig12} presents a comparative analysis of fault tolerance rates across various neuromorphic computing implementations. This comparison includes state of art implementations and our implementation. The chart highlights the percentage of fault tolerance for each study, showcasing how each value in maintaining network integrity under fault conditions. Notably, our implementation demonstrates a significant improvement in fault tolerance, with the lowest rate among the compared studies.

\section{Conclusions}\label{sec:conclusions}
We presented a design methodology for the LIFA model, focusing on integrating astrocytes into neural network models. Our approach involved a core architecture of neurons, synapses, and astrocyte circuits, where each astrocyte encloses multiple neurons for self-repair in case of neuron failure. As a result of this innovative design, the system is much more fault-tolerant, maintaining its efficiency and robustness even under adverse circumstances. An astrocyte-infused network is dynamic and resilient, able to adapt to neuron failures. The LIFA model is mapped onto a fault-tolerant many-core design through a routing methodology that optimizes the network's efficiency and functionality. Based on our rigorous evaluation, the design is both power-efficient and area-efficient, achieving superior fault tolerance. The model we developed outperformed other state-of-the-art implementations by 81.10\% and 18.90\%, respectively. Neuromorphic systems are becoming more resilient and adaptable with the integration of astrocytes.

\bibliographystyle{IEEEtran} 
\bibliography{all} 

\end{document}